\documentclass{article}
\usepackage{pdfpages}   

\usepackage{kr}

\usepackage{xspace}

\usepackage{times}
\usepackage{soul}
\usepackage{url}
\usepackage{xcolor}
\usepackage[hidelinks]{hyperref}
\usepackage[utf8]{inputenc}
\usepackage{graphicx}
\usepackage{amsmath}
\usepackage{amsthm}
\usepackage{booktabs}
\usepackage{etoolbox}
\newcommand{\snJHO}[1]{\textcolor{magenta}{${}\hspace{0pt}{\bullet}$}\marginpar{\textcolor{magenta}{${\leftarrow}\hspace{-3pt}{\bullet}$}\tiny{\textcolor{magenta}{$\:$Jose: #1}}}}

\newcounter{promptcounter}
\newcounter{subpromptcounter}
\def\hide#1{[...]}  

\def\tick{\textcolor{green}{\tiny $\sqrt{\ }$}}
\def\cross{\textcolor{red}{$\times$}}
\def\half{{\tiny \textcolor{orange}{\textonehalf}}}
\def\prompt#1#2{\addtocounter{promptcounter}{1}\setcounter{subpromptcounter}{1}\global\csedef{#1}{\thepromptcounter}\underline{Prompt-\thepromptcounter.\thesubpromptcounter:  }{\textcolor{teal}{\emph{#2}}}}
\def\subprompt#1{\addtocounter{subpromptcounter}{1}\underline{Prompt-\thepromptcounter.\thesubpromptcounter:  }{\textcolor{teal}{\emph{#1}}}}

\def\respT#1{\noindent\underline{Response:  } \textcolor{gray}{{#1}}}
\def\respT#1{\noindent\underline{\chatT-Response:  } \textcolor{gray}{{#1}}}

\def\chat{ChatGPT}  
\def\dalle{DALL-E}
\def\chatT{C3.5}
\def\bard{Bard}
\def\chatPT{P3.5}
\def\chatPF{P4}
\def\chatF{C4}

\title{Dialectical language model evaluation:\\ An initial appraisal of the commonsense spatial reasoning abilities of LLMs}

\author{%
 Anthony G Cohn$^1$\and
Jos\'e Hern\'andez-Orallo$^2$
\affiliations
$^1$University of Leeds \& The Alan Turing Institute \& Tongji University\\
$^2$Universitat Polit\`ecnica de Val\`encia \& ValGRAI \& Leverhulme Centre for the Future of Intelligence   \\
\emails
a.g.cohn@leeds.ac.uk, jorallo@upv.es
}

\begin{document}

\maketitle

\begin{abstract}
  Language models 
  have become very popular recently and many claims have been made about their abilities, including for commonsense reasoning. 
  Given the increasingly better results of current language models on previous static benchmarks for commonsense reasoning, we explore an alternative {\em dialectical} evaluation. 
  The goal of this kind of evaluation is not to obtain an aggregate performance value but to find failures and map the boundaries of the system. Dialoguing with the system gives the opportunity to check for consistency and get more reassurance of these boundaries beyond anecdotal evidence. In this paper we conduct some qualitative investigations of this kind of evaluation for the particular case of spatial reasoning (which is a fundamental aspect of commonsense reasoning). We conclude with some suggestions for future work both to improve the capabilities of language models and to systematise this kind of dialectical evaluation. 
\end{abstract}

\section{Introduction}

Reasoning about common sense has been a goal of AI since its earliest days \cite{mccarthy1959programs}, but has proved surprisingly hard to endow computers with.  Whilst many systems have been built which display expert abilities in fields as varied as medicine (e.g. \cite{van1978mycin}), engineering \cite{dimitrova2020ontological}, bioinformatics \cite{tunyasuvunakool2021highly} and even games 
such Go \cite{silver2016mastering}, there has not yet been a system which has truly displayed the commonsense reasoning abilities of a young child.
However, the advent of large language models (LLMs), such as BERT \cite{devlin-etal-2019-bert} and GPT \cite{brown2020language}, which generally display perhaps surprising fluency and an apparent ability to answer questions relating to almost any topic broached, raises the question of whether, or at least to what extent, they can reason about commonsense notions and answer questions involving common sense.
There are a number of commonsense benchmarks, 
such as the Winograd Schema Challenge (WSC) \cite{levesque2012winograd} and, at least according to aggregate measures, recent LLMs can score well \cite{srivastava2022beyond,Liang-22-holistic}. 

However, such success may not demonstrate a true ability to do commonsense reasoning \cite{WSC-defeat} -- the suggested  reasons 
are (1) lax evaluation criteria, (2) artifacts in the dataset and (3) knowledge leakage from the training data. 
Another important reason is the abuse of aggregations over tasks and collections of tasks, in massive benchmarks, such as BIG-Bench \cite{srivastava2022beyond}.
As has been noted, e.g.  \cite{Burnell23}, even if aggregate measures do indicate good expected performance, they do not necessarily reveal the model limitations, since failed examples are overshadowed by many other examples where the model can use some shortcuts or statistical tricks to get the right answer.  
A fifth reason of why this traditional evaluation of common sense reasoning on language models may be limited is because they take the form of a
static setting, 
in many cases as multiple-choice questions (MCQ); this means that the real extent of understanding cannot be tested, and in the case of MCQs the task is considerably easier since the answer is already provided as  one of the options. 
This multiplies the effects of the second and third issues (artifacts and knowledge leakage from the training set). Actually, many existing tasks and benchmarks have already been used verbatim in the training of the model\footnote{This is a very illustrative example of overstating LM capabilities because of this phenomenon: \url{https://garymarcus.substack.com/p/how-not-to-test-gpt-3}.}. We may use canaries on the training data, or build `fresh' items for evaluation, but this is impractical, as new models appear 
very frequently. 
If we assume this contamination exists \cite[Ap.C/D]{gpt4}, then negative results---refutations, in  scientific terminology---are 
of value,
but positive results should not be taken with the same value. 
Because of all these issues, reasoning, and especially commonsense reasoning, may require a more {\em dialectic} approach to evaluation.

But what is common sense? 
Davis \shortcite{davis2023benchmarks}
suggests the following criteria for common sense: (i) is common; (ii)
is ``largely sensible”; (iii)
supports reasoning; (iv)
is integrated into other cognitive abilities (language, vision, etc.) -- 
one never observes it directly, only how it is manifested through language, action, etc.; (v)
is independent of any modality or task; (vi)
has broad scope; (vii)
is distinguished from common knowledge, encyclopaedic and expert knowledge; (viii)
is concerned with generalities rather than individuals; (ix)
is not book learning or explicitly taught in schools; (x)
is separate from purely linguistic or purely perceptual interpretation. 
Davis goes on to present a list of commonsense benchmarks\footnote{Available at \url{http://cs.nyu.edu/~davise/Benchmarks/}.} and analyses these as to whether they meet the criteria above (and some other desirable features for benchmarks) and concludes ``many of the commonsense benchmarks that have been created do not at all respect
these boundaries; most involve substantial amounts of what is clearly common knowledge and many
involve rather obscure encyclopedic knowledge or, more rarely, even expert knowledge''.   Thus the oft-reported success of AI systems for being able to reason about common sense because of their performance on such benchmarks is questionable.


Given the existing commonsense benchmarks and their limitations, as pointed out by Davis, in this paper we investigate a more granular, dialectic way of evaluating commonsense reasoning in language models. We specifically focus on spatial reasoning, because it should be sufficiently detached from linguistic capabilities and other elements of common sense, it may require a model of the world (intuitive physics) that seems to be at odds with the disembodied character of  language models, and also because spatial reasoning in humans may rely on significant 
inductive bias about the world, such as distance, size, orientation, etc., as the result of our evolutionary history. 
 Modern language models allow for a dialectic evaluation by the accumulation of the previous questions and answers as context. This is what we explore in a manual way here, in the tradition of other dialectic evaluation protocols (e.g., interviews or oral exams) that follow  a systematic choice of questions (unlike the Turing test). In particular, in many cases we 
are influenced by\footnote{McCarthy's concern was about the extent to which the formalism in which a problem is represented would have to be adjusted to meet a variant of the original task.  Here we are not concerned with the representation (which is the impenetrable black box of an LLM), but rather the degree to which that representation is able to solve variant tasks.}
 McCarthy's  \emph{elaboration tolerance}: ``A formalism is elaboration tolerant to the extent that it is convenient to modify a set of facts expressed in the formalism to take into
account new phenomena or changed circumstances''  \shortcite{mccarthy-03}. 

 The purpose of this paper is to explore this dialectical way of probing the system, to start to develop a collection of instances to test for commonsense spatial reasoning abilities, and evaluate 
some LLMs
on these.

\subsection{Commonsense spatial reasoning}

Spatial reasoning (and spatio-temporal reasoning) has long been recognised as a core part of commonsense reasoning, for example spatial representation and reasoning is central to the Naive Physics enterprise \cite{hayes1985second,hayes1985ontology}.
A large part of the AI literature on commonsense reasoning has concerned the representation of and reasoning with, qualitative spatial information  \cite{cohn2008qualitative,chen2015survey}. One of the main reasons for this focus on qualitative spatial representations is that commonsense reasoning is not usually concerned with metric information which tends to be the domain of traditional mathematical reasoning.  Rather, qualitative spatial reasoning is concerned with representing relational information, including notions of relative size or distance\footnote{For example, the well known WSC example of the ``The trophy was too big/small to fit in the case. What was too big/small?\label{trophynote}}. There are a  number of aspects of such spatial knowledge which have been investigated \cite{cohn2008qualitative}, including (i) mereotopology, (ii) direction and orientation, (iii) size, (iv) distance and (v) shape.
Frequently, spatial reasoning requires reasoning about more than one of these aspects simultaneously.  

There are certainly spatial instances in the various  benchmarks purporting to test commonsense reasoning, some of them in multimodal situations \cite{liu2022visual}, but others in purely textual form. 
We could use some instances from existing benchmarks such as from the WSC, but these are vulnerable to statistical patterns and other tricks \cite{WSC-defeat}, and are limited to static non-dialectic interactions. Also, there is the contamination risk that the LLM may have been trained on them, making good results not indicative of a generic capability. We will build on some of them in section \ref{sec:wsc} though, and some other sources in section \ref{sec:page}, since the dialectic character of the evaluation allows for elaboration on the reasons or variations, to check if the model has or does not have the ability.  

\section{An evaluation of the spatial reasoning abilities of some LLMs}

In order to exploit the dialectical possibilities for evaluating commonsense reasoning and the use of context (previous interaction in the session), we will use conversational foundation models. 
One of these is  \chat\ 
as it is (a) recent, (b) easily accessible and (c) has been the subject of an extraordinary amount of attention in the scientific and popular media.  We evaluate a number of versions of \chat\   using GPT-3.5turbo and also using GPT-4 on chat.openai.com; and using  3.5turbo and also using GPT-4 on platform.openai.com/playground. Finally we also evaluate \bard, based on LaMDA \cite{thoppilan2022lamda}. For convenience below, we refer to these five LLMs as \chatT, \chatF,  \chatPT, \chatPF\  and \bard\  respectively.


\chat\ and \bard  can certainly provide very fluent and sometimes perhaps surprisingly sophisticated outputs, e.g. when asked to write something in a particular style. 
However, to what extent does such fluent text output display  commonsense reasoning abilities and in particular about space? This is the question we now turn to. In this paper we cannot hope to exhaustively answer this question; rather, we sample the space, guided by some existing challenges and by basic competencies we might expect to be displayed. In what follows, we include the results for \chat\  using GPT-3.5turbo (\chatT), because it shows a greater number of inadequate or incorrect responses, and illustrates the dialectic approach best.
Actually, for those responses that are correct we simple include a \tick\ symbol omitting the answer.
\footnote{All prompts are numbered in the form x.y for convenience in referring to them later; x refers to the current conversation, and y to the sequence of prompts within it. Conversational LLMs  often give quite long responses, with the final part summarising (or not!) the earlier part of the response. In the interests of space economy, in general we do not give the responses here where they are acceptable, and in the cases where the LLM makes an error  we only give the key part(s) of the response, with   sections indicated by ellipsis [...].  The full prompt-response pairs can be found 
in the appendix to this paper.
The summary of results for \chatT\  can be found in Table \ref{results-all}, while the comparison for all LLMs can be found in Table \ref{results-all}. }
In some of the conversations reported here, asking later prompts in a conversation separately in a new   conversation sometimes will give an improved result. However, keeping all the prompts in a single chat session as  a single dialogue unveils some important  and revealing mistakes, and this is the case we report.


\subsection{Basic spatial relations}
\subsubsection{Parthood}
Parthood is perhaps the most basic of all spatial relationships.  Mereology \cite{simons2000parts} has been widely studied in the literature and forms the basis of much of qualitative spatial reasoning. Let us start with some dialogues. 
\prompt{spatialpart}{If x is part of y and y is part of z what is the spatial relationship between x and z?}
\respT{
If x is part of y and y is part of z, it means that x is a subcomponent of y, and y is a subcomponent of z. \hide{In other words, x is a part of y, and y is a part of z.}
The spatial relationship between x and z depends on the specific context of the situation. It is possible that x and z are physically connected or adjacent to each other, but it is also possible that they are spatially separate from each other.
For example, if x is a tire and y is a car, and z is a garage, then x is part of y and y is part of z. However, x and z are not necessarily spatially related to each other. The tire may be physically located in the garage, but it could also be in the trunk of the car, or even outside the garage.
\hide{In general, the spatial relationship between x and z cannot be determined solely based on the fact that x is part of y and y is part of z. Additional information about the context of the situation is necessary to determine the spatial relationship between x and z.}
} The response here is not good; the first sentence just recapitulates the question and then a very odd example ensues.  Most would agree that a tyre is part of car, but not that a car is part of garage as seems to be implied here (it might be \emph{in} garage though -- see below). If something is a subcomponent of something else one would always regard it as spatially related and indeed a spatial part. 

``Part of'' and ``in''  tend to be used interchangeably so we explore the response response in this case.
\subprompt{If  u is in v and v is in w, what is the spatial relationship between u and w?} \respT{
If u is in v and v is in w, it means that u is a subcomponent of v, and v is a subcomponent of w. \hide{ In other words, u is a part of v, and v is a part of w.}
The spatial relationship between u and w depends on the specific context of the situation. However, in general, u and w are not necessarily spatially related to each other, just like in the previous question.
For example, if u is a cell, v is an organ, and w is a body, then u is in v and v is in w. However, u and w are not necessarily spatially related to each other. The cell could be located in the organ, but the organ could be located in a different part of the body.
\hide{In general, the spatial relationship between u and w cannot be determined solely based on the fact that u is in v and v is in w. Additional information about the context of the situation is necessary to determine the spatial relationship between u and w.}} As with prompt \spatialpart.1,  \chatT\  presents a confusing example, claiming in a contradictory manner that a cell could be in an organ which is in a different part of the body to where the cell is! 
We then tried using ``contains'' rather than ``part of''.
\subprompt{If f contains g and g contains h then what is the spatial relationship between f and h?}
\respT{
\hide{If f contains g and g contains h, it means that h is a subcomponent of g, and g is a subcomponent of f. In other words, h is a part of g, and g is a part of f.}
Since h is a subcomponent of g, and g is a subcomponent of f, it follows that h is also a subcomponent of f. In other words, h is contained within f.
Therefore, the spatial relationship between f and h is that h is located within f. In other words, h is a part of f, and may be physically located within f, depending on the specific context of the situation.}  The response here is good except for the unwanted hedge in the last sentence. 

Finally we explicitly state in the prompt that the relationship between the entities involved are spatial relations rather than leaving this implicit (to see if this helps with the strange example generation). 
\subprompt{If b is a spatial part of c and c is a spatial part of d, then what is the spatial relationship between b and d?}
\respT{\tick}.

\subsubsection{Rotation}
Reasoning about rotation is a commonly used test of spatial reasoning abilities. In the absence of diagrams, here we use letters which have a known shape.
\prompt{rotationM}{If I rotate a letter M through 180 degrees what letter is it most similar to?}
However, \chatT\  performs badly on a different letter rotation test:
\subprompt{
If I rotate a capital Z by 90 degrees what letter is it most similar to?}
\respT{If you rotate a capital ``Z" by 90 degrees clockwise, it would look most similar to the letter ``L" (in uppercase form).
The capital ``Z" is a diagonal letter that consists of two perpendicular lines. When you rotate it by 90 degrees, the two lines become horizontal, with the top line now on the left and the bottom line now on the right. This shape closely resembles the letter ``L", which is also a horizontal letter consisting of a single line on the left and a shorter line on the right.\hide{ So, if you rotate a capital ``Z" by 90 degrees clockwise, it would look most similar to a capital ``L".}}  \chatT\  incorrectly describes the shape of a Z and then gets the wrong answer.  However if asked about rotating an N, it gets the right answer, in spite of incorrectly describing the shape of both an N and Z.
\subprompt{If I rotate a capital ``N" by 90 degrees then what letter is it most similar to?}
\respT{If you rotate a capital ``N" by 90 degrees clockwise, it would look most similar to the letter ``Z" (in uppercase form).
The capital ``N" consists of two diagonal lines that intersect at a right angle, with the longer line pointing downwards to the right. When you rotate it by 90 degrees, the longer line will point to the left instead of downwards, and the shorter line will point upwards instead of to the left. This shape closely resembles the letter ``Z", which is also a diagonal letter consisting of two lines that intersect at a right angle, with the longer line pointing downwards to the right. \hide{So, if you rotate a capital ``N" by 90 degrees clockwise, it would look most similar to a capital ``Z".}} 

\subsubsection{Directions}
We 
restrict ourselves here to
 asking about spatial points since directions between areal objects can be more complex. \prompt{direction}{If a point b is to the left of a point c, and a point d is to the right of a point c, then what is the spatial relationship between b and d?}
\respT{
If point b is to the left of point c and point d is to the right of point c, then point b and point d are on opposite sides of point c. In other words, point b is on the left-hand side of a line passing through points c and d, while point d is on the right-hand side of the same line.
Therefore, the spatial relationship between b and d is that they are on opposite sides of the line passing through c and d.}  This looks good but is not completely clear about the spatial relationship between b and d, only the relationship between of these and the constructed line. Unfortunately \chatT\   is not able to answer this question.
\subprompt{is b to the left or right of d?}
\respT{Without additional information about the positions of points b and d relative to each other, it's impossible to determine whether b is to the left or right of d.
All we know is that b and d are on opposite sides of the line passing through c and d. But their positions relative to each other could be anywhere on their respective sides of that line.}
Given this, perhaps unsurprisingly, \chatT\  is not able to reason about left/right and above/below simultaneously.
\prompt{imageplane}{In an image plane, if point e is directly to the left of point f, and point g is directly above point f, then what is the spatial relationship between e and g in that image plane?}
\respT{Without more information, we cannot determine the spatial relationship between e and g with certainty.
If e and g are on the same horizontal line in the image plane, then e and g would be neither directly above nor below each other. In this case, e would be to the left of f, and g would be above f, but e and g would not be directly related to each other in terms of their horizontal or vertical positions.
However, if e and g are on different horizontal lines in the image plane, then they could have a different spatial relationship. For example, if e is on a higher horizontal line than g, then e would be to the left and above g. On the other hand, if g is on a higher horizontal line than e, then e would be to the left and below g.
\hide{Therefore, we need more information about the positions of e, f, and g in the image plane to determine their spatial relationship.}}


\subsection{Size, shape and location}\label{sec:wsc}

\subsubsection{Circle size}
Reasoning about relative size is another important aspect of spatial common sense. 
\prompt{circlesize}{If a circle b is larger than a circle c, is it possible to move b so that it is entirely contained by c?}
\respT{Yes, it is possible to move a larger circle (b) so that it is entirely contained by a smaller circle (c).
To do this, you would need to move the center of circle b towards the center of circle c until the edge of circle b touches the edge of circle c. As long as the radius of circle b is smaller than the distance between the centers of the two circles, it is possible to move circle b in this way so that it is entirely contained by circle c.
It's important to note that in order for circle b to be entirely contained by circle c, the radius of circle b must be less than or equal to the difference between the radii of circle c and circle b.} 
\chatT\  is wrong in its answer and also, in the last sentence, in its specification of what it would take for it to be possible.

\subsubsection{Trophy}
For the WSC trophy/case example already mentioned in footnote \ref{trophynote},
\chatT\  fails to reason correctly, despite the example being widely available on the web (though usually without the correct answer given).  \prompt{trophy}{In the sentence ``The trophy wouldn't fit in the case because it was too small.", what does the ``it" refer to?} \respT{
In the sentence ``The trophy wouldn't fit in the case because it was too small.", the pronoun ``it" refers to the trophy.  }  The answer is wrong so we asked for a justification:  \subprompt{Why do you say that?
} \respT{
I say that because the subject of the sentence is ``the trophy" and the pronoun ``it" is used later on in the sentence as a pronoun to refer back to the subject ``trophy".
\hide{This is a common way that pronouns are used in English to avoid repeating the same noun over and over. In this case, ``it" serves as a stand-in for ``the trophy" and refers to the same thing.}} \chatT\  has justified its answer purely in terms of linguistic rather than semantic knowledge, so we challenged it again: \subprompt{
Why doesn't ``it" refer to the case?} \respT{
``It" doesn't refer to the case because the subject of the sentence is ``the trophy", not ``the case". The sentence is talking about the trophy not fitting into the case, so the focus is on the trophy and not the case. Additionally, the sentence says that ``it was too small", which would make more sense if ``it" referred to the trophy, since the size of the trophy would be more relevant to whether or not it could fit in the case. If ``it" referred to the case, the sentence would be saying that the case was too small, which is not what the sentence is trying to convey.
} \chatT\  displays ever more confusion, 
and doesn't appreciate that the size of the case could also be relevant. The confusion continues when the other half of the WSC is posed: \subprompt{In the sentence ``The trophy wouldn't fit in the case because it was too big.", what does the ``it" refer to?
} \respT{\hide{In the sentence ``The trophy wouldn't fit in the case because it was too big.",}  
the pronoun ``it" cannot logically refer to the trophy, as the sentence says the trophy is too big to fit in the case. Therefore, if the sentence is correct, ``it" must refer to the case. However, the sentence contains a contradiction, because the first part of the sentence implies that the trophy is too large to fit in the case, but the second part says that the case is too big, which suggests that the trophy would actually fit.
If we assume that the intended meaning of the sentence is that the trophy is too big to fit in the case, then the correct sentence would be ``The trophy wouldn't fit in the case because it was too small''.
} \chatT\   contradicts itself in the first sentence and then goes on to insist that the ``it'' must refer to the case before finally wanting to incorrectly rephrase the sentence. When challenged 
\subprompt{Why can't the ``it" refer to the trophy?} \chatT\  does apologise for getting wrong but then gives an incorrect reason for the change of mind:
\respT{
\hide{You are correct, if the sentence is ``The trophy wouldn't fit in the case because it was too big," then ``it" could logically refer to the trophy. I apologize for my previous response in which I misread the sentence.}
In this case, the use of the word ``it" indicates that the speaker is referring to the trophy, as the trophy is the last noun mentioned before the pronoun ``it." 
\hide{The word ``it" serves as a pronoun that refers back to the noun ``trophy" to avoid repeating the same noun twice in the sentence.
So, in the corrected sentence ``The trophy wouldn't fit in the case because it was too big," the pronoun ``it" refers to the trophy.} }
whilst also showing it cannot reason correctly  about spatial adjacency.

\subsubsection{Bookcase}
We invented the following analogous WSC to see how consistent the problem is. \prompt{bookcase}{The book couldn't stand upright in the bookcase  because it was too small.  What does ``it" refer to?} \respT{
``It" in the given sentence refers to the book.
\hide{The sentence means that the book couldn't stand upright in the bookcase}
because the book was too small to fit properly.} As it is wrong, we ask why. \subprompt{
Why do you say that?} \respT{
\hide{I say that because the sentence is structured such that the subject is ``the book", and the verb phrase is ``couldn't stand upright in the bookcase".}
The reason for the book's inability to stand upright is given by the dependent clause ``because it was too small." 
\hide{In this dependent clause, }
``it" is used as a pronoun to refer back to the subject of the sentence, which is ``the book". 
\hide{Therefore, ``it" in the sentence refers to the book.}}
\subprompt{
Does a pronoun always refer to the subject of the sentence?} 
\respT{\tick}.
\subprompt{Surely the book wouldn't fit properly only if it was too big, not too small?} \respT{
You are correct. \hide{ I apologize for my previous response.} The correct interpretation of the sentence is that ``the book couldn't stand upright in the bookcase because it was too big." When a book is too big, it cannot fit properly in a bookcase, and as a result, it cannot stand upright.
\hide{I apologize for any confusion my previous response may have caused, and I appreciate you bringing this to my attention. Thank you.}}  
So finally \chatT\  does get the right answer and 
reason. But when re-asked the original question, \chatT\  gets  it wrong still (in spite of being within the same \chatT\  session): \subprompt{
So if I say ``The book couldn't stand upright in the bookcase because it was too small", what does the ``it" refer to?} \respT{
In the sentence ``The book couldn't stand upright in the bookcase because it was too small," the pronoun ``it" refers to the book.
The sentence suggests that the book is too small to fit properly in the bookcase, which is causing it to be unable to stand upright. Therefore, the word ``it" in the sentence refers to the book, which is the subject of the sentence.} At least the justification is not made on linguistic grounds but rather on putative semantic ones, but clearly fails to understand commonsense spatiality here\footnote{Perhaps an argument could be made that the book couldn't stand  upright because it is too small to be stable in an upright position, but \chatT\ claims that it is  not fitting.}.    When given the complementary sentence, 
however, \chatT\  correctly gives the referent of ``it'':
\subprompt{
In the sentence ``The book couldn't stand upright in the bookcase because it was too big", what does the ``it" refer to?} \respT{\tick}.

\subsubsection{School bag}
Another spatial WSC is this one: \prompt{schoolbag}{Tom threw his school bag down to Ray after he reached the top  of the stairs. Who reached the top of the stairs? } \respT{
Tom reached the top of the stairs.} The answer is correct, though based on the previous examples we may suspect that it has not reasoned semantically but rather because Tom is the subject of the sentence, which seems to be confirmed in the case of the complementary schema: \subprompt{
Tom threw his school
bag down to Ray after he reached the bottom of the stairs.
Who reached the bottom of the stairs? } \respT{
The sentence you provided is contradictory, as it states that Tom threw his school bag down to Ray after he (Tom) reached the bottom of the stairs. If Tom reached the bottom of the stairs, he would not be in a position to throw his school bag up to Ray, who would presumably be at the top of the stairs. It's possible that you meant to say ``after he (Tom) reached the top of the stairs," in which case the answer would be Ray reached the bottom of the stairs. \hide{Please let me know if this is the case, or if you have any other questions.}}  It is also not clear why Tom couldn't throw the bag up to Ray (unless it was particularly heavy), though of course the question was about throwing the bag down not up. \chatT\  does finally give the right answer when given a further
\subprompt{
No it's not contradictory if you interpret ``he" as referring to the correct person.} 
\respT{
I apologize for misunderstanding your question earlier. In the sentence ``Tom threw his school bag down to Ray after he reached the bottom of the stairs," the pronoun ``he" refers to Ray. Therefore, Ray is the person who reached the bottom of the stairs.}  
We can test its understanding of property transfer: \subprompt{So who has the school bag now?} \chatT\  is able to respond correctly: 
\respT{
Based on the sentence ``Tom threw his school bag down to Ray after he reached the bottom of the stairs," we can infer that Ray has the school bag now, as Tom threw it down to him.} 
Are the systems  able to reason correctly on the same sentence with substituted names? 
\subprompt{
John threw his school bag down to Frank after he reached the bottom of the stairs. Who reached the bottom of the stairs? } 
Unfortunately the answer is in the negative: 
\respT{
In the sentence ``John threw his school bag down to Frank after he reached the bottom of the stairs," the pronoun ``he" refers to John. Therefore, John is the person who reached the bottom of the stairs.} 
\subsection{Affordances and object interaction}\label{sec:page}
The Commmonsense Problem page\footnote{\url{http://www-formal.stanford.edu/leora/commonsense/}.} (CPP) contains a selection of problems contributed over a period of some years by members of the community which represent ``challenge problems" for commonsense reasoning.  Not all of these involve spatial reasoning; in this section we look at the performance of the systems  on some of those that do.
\subsubsection{Stakes in a Garden} 
\prompt{plants}{A gardener who has valuable plants with long delicate stems protects them against the wind by staking them; that is, by plunging a stake into the ground near them and attaching the plants to the stake with string. Why will this help?}\footnote{The specification on the CPP asks to characterise the following physical operation rather than asking ``why". The variants we use are also slightly rephrased too.}
\respT{Staking delicate plants with long stems can help protect them from wind damage in several ways. 
Firstly, wind can cause the plant to sway or bend, and if the stems are long and delicate, they may break under the stress. Staking provides support for the plant, keeping it upright and preventing it from bending or breaking. 
Secondly, wind can cause soil erosion around the base of the plant, which can destabilize it and cause it to topple over. Staking can help anchor the plant in place, preventing it from being uprooted by the wind. 
Lastly, staking can help promote healthy growth in the plant. \hide{When a plant is constantly swaying in the wind, it may put more energy into growing thicker stems to support itself rather than producing leaves or flowers. By providing support through staking, the plant can focus its energy on producing healthy foliage and blooms.}  
Overall, staking delicate plants with long stems can help protect them from wind damage, promote healthy growth, and ensure they continue to thrive.}  The response is good .  CPP follows McCarthy's elaboration tolerance with a number of variants some of which were also put to the systems. 
\subprompt{Would it work if the length of the string was shorter than the distance between the plant and the stake?} \respT{If the length of the string used to tie the plant to the stake is shorter than the distance between the plant and the stake, it may not provide adequate support to the plant. The string may be too tight, which can cause the stem to bend or break, or it may not be able to support the weight of the plant. 
In order to properly stake a plant, the string should be long enough to wrap around the stem several times and then be tied to the stake with a knot or a twist tie. This ensures that the string is not too tight and can provide support to the stem without damaging it. 
If the distance between the plant and the stake is too great, it may be necessary to use additional stakes or to reposition the stake closer to the plant. \hide{Proper staking is essential to prevent wind damage and promote healthy growth in delicate plants with long stems.}}  The response here does not capture the essential point that a short string could not provide \emph{any} support, though at least it does suggest a longer string might be better. 
Conversely for this \subprompt{Would it work if the length of the string was much longer than the distance between the plant and the stake?} \chatT\ gives a good response, even noting the risk of entanglement: \respT{If the length of the string used to tie the plant to the stake is much longer than the distance between the plant and the stake, it may not be as effective in supporting the plant. The excess string may cause the plant to sway or move around more than necessary, which could potentially damage the stem or roots. 
Additionally, the excess string may create a tripping hazard or become tangled with nearby plants or garden tools. 
Ideally, the string used to tie the plant to the stake should be just long enough to wrap around the stem several times and then be tied to the stake with a knot or a twist tie. This will provide the necessary support to the plant without causing any unnecessary movement or creating any hazards in the garden. 
In some cases, using a longer string may be necessary if the plant is particularly tall or the stake needs to be placed far away from the plant for some reason. However, it's important to use the right length of string for each specific situation to ensure that the plant is properly supported and that the garden remains safe and tidy.}  
\subprompt{Could I twist the ends of the string together instead of making a knot?} \respT{Yes, twisting the ends of the string together can be an effective way to tie a plant to a stake. This is known as a ``twist tie" and it can be a convenient and secure method for attaching the string to the stake. 
\hide{To create a twist tie, you will need to wrap the string around the plant stem and then twist the ends of the string together until they are tight. Once the string is securely in place, you can wrap the twisted end around the stake and twist it again to secure it.} 
Twist ties can be particularly useful for delicate plants with soft stems, as they can be tied tightly without damaging the plant. They can also be easily adjusted or removed as needed, which makes them a convenient option for plants that may require frequent staking. 
Overall, whether you choose to tie the string in a knot or use a twist tie, the most important thing is to ensure that the plant is properly supported and that the tie is not too tight or too loose. This will help prevent damage to the plant and promote healthy growth.}  Here \chatT\  doesn't show any understanding of what twisting the ends of the string means and also implies that it might damage the plant which would not be the case.  The end of the response concerning the knot not being too tight does not show real understanding of how the string should be used (i.e. to keep the plant close to the stake rather than tying it to it).
\subsubsection{The Handle Problem}
 This problem concerns trying to formalise what a handle is and what could be used as a handle. \prompt{handle}{Give a general purpose characterisation of what constitutes a handle, in the ordinary sense of door-handle or drawer-handle, which is sufficient to enable one to infer from a qualitative description of the shape of a part of an object whether or not it can be a handle for that object.} \respT{In the ordinary sense, a handle is a structural component of an object that is intended to be grasped by the user to facilitate movement or manipulation of the object. A handle is typically a protruding part that provides a surface for the user to grip onto, and it is designed to be ergonomic, easy to grasp, and durable.
Based on this general purpose characterization of a handle, one can infer whether or not a part of an object can be a handle by considering the following factors:
    Protrusion: A handle typically protrudes from the surface of the object to provide a surface for the user to grip onto. \hide{If the part of the object in question does not protrude, it may not be suitable for use as a handle.}
    Ergonomics: A handle should be designed to fit comfortably in the user's hand, with a surface that is easy to grip onto. \hide{If the shape of the part of the object does not appear to be ergonomically designed for this purpose, it may not be a suitable handle.}
    Durability: A handle should be durable enough to withstand repeated use without breaking or wearing down. \hide{If the part of the object does not appear to be sturdy enough to serve as a handle, it may not be a suitable option.
Overall, a handle is a structural component that is designed to be grasped by the user to facilitate movement or manipulation of an object. By considering the protrusion, ergonomics, and durability of a part of an object, one can infer whether or not it can function as a handle.}}
The response is a good characterisation, though not really surprising given the large numbers of definitions of handles likely to be in its training data.
\subprompt{
Could I use a blunt conical projection as a handle for a door or drawer?} \respT{Yes, a blunt conical projection can be used as a handle for a door or drawer, depending on its design and size. A conical projection can provide a comfortable grip for the user and can be easy to operate.
The size and shape of the conical projection should be designed with ergonomics in mind, ensuring that it fits comfortably in the user's hand and is easy to grasp. Additionally, the conical projection should be durable enough to withstand repeated use without breaking or wearing down.
It's worth noting that a blunt conical projection may not be the most common type of handle for doors or drawers, as other shapes such as cylindrical or rectangular handles are more commonly used. \hide{However, if designed properly, a blunt conical projection can function effectively as a handle.}}  This is clearly a failure of commonsense reasoning, although it does note it's not a common shape -- the response fails to realise that this really would not be very effective as a handle. So we checked which orientation \chatT\  thought the handle was in, and responded that it is indeed tapering away from the drawer: 
\subprompt{Is the base of this conical handle next to or away from the drawer?} \respT{The orientation of the conical handle base in relation to the drawer will depend on the specific design of the handle. However, in most cases, the base of the conical handle would be attached to the surface of the drawer or door, with the tapering section of the handle extending away from the drawer or door.
This orientation allows the user to grasp the tapered section of the handle and pull it outward, which in turn opens the drawer or door. The base of the handle provides a stable attachment point for the handle, ensuring that it remains securely in place when in use.
\hide{Of course, there are many variations on this basic design, and the specific orientation of the handle base may vary depending on factors such as the size and shape of the drawer or door, the intended use of the handle, and the preferences of the designer or manufacturer. However, in general, a conical handle for a drawer or door would be oriented with the base attached to the surface of the drawer or door and the tapering section extending outward.}}  So \chatT\  does believe that the base is next to the door/drawer. One strong point of this response is the comment that attaching the handle this way round provides a secure attachment point (not literally a point of course!) which displays some spatio-mechanical knowledge. A negative point is that this orientation does not allow a good grasp.
So we challenged \chatT\  about the orientation: \subprompt{Wouldn't it be more ergonomic to have the base away from the drawer?} \respT{Having the base of the conical handle away from the drawer or door could provide a different grip orientation and could be more ergonomic for certain users or situations. This design could allow the user to wrap their fingers around the base of the handle, which could provide a more comfortable and secure grip, especially if the handle is being used to lift a heavy object.
However, this design could also make it more difficult for the user to operate the drawer or door, as they would need to push the handle inward instead of pulling it outward. \hide{This could be especially problematic if the drawer or door is heavy or difficult to move.
Ultimately, the choice of handle design will depend on a variety of factors, including the size and shape of the object being handled, the intended use of the handle, and the preferences and needs of the user.} A handle with the base attached to the surface of the drawer or door and the tapering section extending outward is a common and effective design, but there may be situations where an alternative design, such as one with the base away from the drawer, could be more appropriate.}  The first  part is a good response, but the middle section is bizarre and clearly fails a lack of understanding about how opening a drawer works. It also displays a lack of common knowledge in the final part where it claims that this orientation is common.

\subsubsection{Reasoning about linked chains}
Another spatial problem in the CPP concerns chains\footnote{The formulation of the initial prompt is changed to try to make it clear what kind of chain is envisaged (open loop rather than a bicycle type chain for instance) -- in a prior conversation \chatT\  claimed that chains had pins,  rivets and hinges.}: 
\prompt{chain}{Formally characterize the structure of a metallic linked chain. What will happen if one picks up one end of such a chain and walks away? By a linked chain I mean a chain which is constructed as a series of interconnected loops of metal and no other parts, such as is used as an anchor chain in a ship.}
\respT{A metallic linked chain is a structure composed of interconnected metal loops that form a continuous chain. Each loop in the chain is linked to the adjacent loop through a mechanical joint, which provides flexibility and allows the chain to bend and conform to different shapes.
When one end of the linked chain is picked up and pulled away, the rest of the chain will follow due to the interlinked nature of the loops. As the chain is pulled, the links will stretch and become elongated, causing the chain to become thinner and longer. The chain will also experience tension forces along its length, which will increase as the pulling force is applied. If the force applied is too great, the chain may break at its weakest point, or one of the links may become dislodged from the rest of the chain. 
\hide{Therefore, it is important to consider the maximum load capacity of the chain and the force that can be applied without causing damage before attempting to pull it.}}
This response starts off well (predicting bending and confirming a following behaviour) but then becomes flawed: one wouldn't expect such a metal chain to stretch (except by virtue of the links always being tight up against its neighbours) and it would not become thinner (as a string of Plasticine would). The conversation continued:
\subprompt{If the loops of this chain are large, is it possible to move the last link in the chain at all without any other link moving?}
\respT{If the loops of the chain are large, it may be possible to move the last link in the chain without any other link moving, but it depends on the specific geometry and construction of the chain.
\hide{In general, a metallic linked chain is designed to be flexible and able to bend in multiple directions, which means that there will be some degree of movement or "play" between adjacent links. This play is what allows the chain to conform to different shapes and absorb shock loads without breaking.}
If the last link in the chain is located near the end of the chain and there is enough slack or play between adjacent links, it may be possible to move the last link without any other link moving. However, if the last link is located closer to the middle of the chain, \hide{or if the chain is under tension, it may be difficult or impossible to move the last link without also moving adjacent links.
In general, the larger the loops in the chain and the more slack there is between adjacent links, the easier it will be to move the last link without any other link moving. However, the specific details of the chain's geometry and construction will ultimately determine whether  this is possible or not.
} }
This is mostly a good response but \chatT\  is clearly confused about where the end of chain is, saying it might be in the middle! We then probed its understanding of what it means to be the last link in a chain. 
\subprompt{How many other links is the last link in a chain connected to?}
\respT{The number of other links that the last link in a metallic linked chain is connected to will depend on the specific design and construction of the chain.
In general, a metallic linked chain is designed to be a continuous loop, with each link connected to the adjacent link through a mechanical joint. The last link in the chain will be connected to the second-to-last link, which in turn will be connected to the third-to-last link, and so on, until all of the links in the chain are connected in a continuous loop.
\hide{However, the last link in the chain may also be connected to other components or structures, depending on how the chain is being used. For example, in an anchor chain used on a ship, the last link may be connected to the ship's anchor or to a shackle that attaches the chain to the anchor. Similarly, in a lifting chain used to lift heavy loads, the last link may be connected to a hook or other lifting device.}
Therefore, while the last link in a metallic linked chain is typically connected to one or more adjacent links, the number of other links it is connected to may vary depending on the specific application and use case of the chain.}
There are several odd things here:  \chatT\   seems to think a chain is necessarily a loop rather than being just \emph{made} of loops.  But in that case there would not be a last link!  And it explicitly suggests   that the last link is connected to more than one other link.  So we pressed \chatT\  further. 
\subprompt{Earlier you said that the last link could be in the middle of the chain. Can you explain further?}
\respT{I apologize for the confusion in my previous response. In a continuous loop metallic linked chain, the last link will always be connected to the second-to-last link, and the second-to-last link will be connected to the third-to-last link, and so on. Therefore, it is not possible for the last link to be in the middle of the chain.
However, it is possible for the chain to have a break in the loop, in which case the last link may not be connected to any other links. For example, if a section of the chain is removed or cut, the last link on one side of the break will no longer be connected to the second-to-last link on the other side of the break.
If a break occurs in the chain, the links on either side of the break will no longer be connected in a continuous loop, and the chain will no longer be able to function as intended. \hide{In this case, the broken section of the chain would need to be repaired or replaced in order to restore the chain to its original function.}
}
This is good except for \chatT's insistence that chains must be a circular and cannot function as intended if not a loop.
\subprompt{If a chain is used a ship's anchor chain, is the chain a continuous loop?}
\respT{
Yes, a chain used as a ship's anchor chain is typically a continuous loop metallic linked chain. The chain is connected to the ship at one end and to the anchor at the other end, forming a continuous loop that allows the anchor to be lowered and raised as needed.
\hide{The chain used as a ship's anchor chain is designed to be strong, durable, and resistant to corrosion from exposure to saltwater and other harsh marine environments. It is typically made from high-strength steel or other materials with similar properties, and each link in the chain is designed to interlock with the adjacent link to form a continuous loop.
In addition to providing a secure anchor point for the ship, the chain also helps to absorb the shock loads and stresses that can be generated by wave action and other external forces. By distributing these loads and stresses throughout the length of the chain, rather than concentrating them at a single point, the chain helps to protect the ship's hull and other components from damage.
Overall, a continuous loop metallic linked chain is an essential component of a ship's anchoring system, providing a strong and reliable connection between the ship and the anchor that allows the ship to remain safely moored in a variety of sea conditions.
}}
This is not only partly wrong but also self contradictory as a loop does not have ends!
We then pressed \chatT\  specifically about the nature of the chain. 
\subprompt{
When you talk of a ``continuous loop metallic linked chain", do you mean that the components of the chain are continuous loops, or that the chain itself forms a continuous loop?}
\respT{
When I refer to a ``continuous loop metallic linked chain," I mean that the chain itself forms a continuous loop, not necessarily that the components of the chain are continuous loops.
A continuous loop metallic linked chain is a chain in which each link is connected to the adjacent link to form a continuous loop. The chain can be made up of individual links that are not continuous loops themselves, but when connected, they form a loop that is continuous and unbroken.
\hide{The links in a continuous loop metallic linked chain are typically designed to interlock with each other, forming a strong and secure connection that can withstand the stresses and loads placed on the chain.
The chain may be made up of links that are all the same size and shape, or it may have different types of links that are used for specific purposes, such as anchor chain, lifting chain, or chain used in a conveyor system.}
In general, a continuous loop metallic linked chain is designed to be flexible and able to bend in multiple directions, which allows it to conform to different shapes and absorb shock loads without breaking. This flexibility also allows the chain to be wound up and stored in a compact space when not in use.
\hide{Overall, the continuous loop design of a metallic linked chain provides strength, durability, and versatility, making it a useful component in a wide range of industrial, commercial, and maritime applications.}}
\chatT\  is not able to consistently explain the nature of a chain and contains several inaccuracies.

\subsection{Object permanence} 

In this section we present some other problems specifically invented in this paper for the purposes of probing the systems' understanding of object permanence.
\subsubsection{Where are the car keys?}
This problem refers to object permanence when 
an 
agent is moving (egocentric object permanence). 
\prompt{keys}{John and Mary are having breakfast together and Mary puts the car keys on the table.  Mary then goes out for a walk.   Whilst she is out, John puts the keys in his pocket.  Mary then returns.  Where does Mary expect to find the keys?}
\respT{
Mary likely expects to find the keys where she left them, which is on the table. Since she did not give any indication to John that he could take the keys or move them, she would assume that they are still in the same place where she left them.
If she doesn't find the keys on the table, she may ask John about it, but it's also possible that John may return the keys to the table before she realizes they are missing. In any case, it would be important for John to communicate with Mary if he moved the keys, so as to avoid any confusion or inconvenience.}  This answer is good  -- just what one would hope for. However if we then ask: \subprompt{
If the car wasn't in the drive when Mary returned from the walk, and John was out, then where would she expect the keys to be?} then we run into problems: 
\respT{
If the car wasn't in the drive when Mary returned from the walk and John was not there, Mary would likely check the usual places where they keep the keys, such as a key holder, a drawer, a hook, or a purse. If Mary could not find the keys in any of the places, she might have to wait for John to return or call or message him to ask where the car is and where the keys are.
If John had taken the keys with him and was not reachable, Mary might have to find alternative transportation or wait for John to return. If Mary had a spare set of keys, she might be able to use those to drive the car.
\hide{In any case, if John took the keys without informing Mary, he should let her know the whereabouts of the car and keys as soon as possible, so that she is aware of their location and can make any necessary arrangements. Communication is important to avoid misunderstandings and inconvenience.}} \chatT\   has completely failed to connect the change of spatial position of the car with the change of spatial position of the keys.  Worse, it goes on to suggest that she might use some spare keys to drive the car which is not in the drive any more!

\subsubsection{Cars behind the sofa}
Now let us try an example where it's the object moving (allocentric object permanence). 
\prompt{cars}{You are the only person in a room and see a green toy car on the left, a sofa in front of you and a door on the right. A car starts moving from left to right and you can see how it approaches the sofa. Then you don't see the car for a while. Finally a car appears again approaching the door. What happened while you couldn't see any car?}
\respT{\hide{As an AI language model, I don't have visual perception, so I can't see anything. However, }based on the information you've provided, it seems that the green toy car passed behind the sofa while you couldn't see it. This suggests that there may have been an obstruction blocking your view of the car for a period of time, such as the sofa.}
This is very good. It seems \chatT\ understands object permanence with occlusion. But let us probe a bit further.
\subprompt{The car that appeared on the right was red. Can you explain what happened?}
\respT{Certainly! Based on the information you've provided, it seems that there were two toy cars in the room - a green one and a red one. The green car initially started moving from the left side of the room towards the sofa in front of you, and then passed behind the sofa while you couldn't see it. After a while, the red car appeared, moving towards the door on the right side of the room.
It's possible that the red car was already present in the room, but not initially visible to you due to your line of sight being blocked by the green car or another object in the room. Alternatively, the red car may have entered the room at some point while you were not observing it. Without more information, it's difficult to say for sure what exactly happened.}
It starts well, but the final part shows important problems, such as thinking that the red car was occluded by the green car, or the comment that the red car could have come from the door, when we said it was approaching the door.

\section{Discussion}

As we have seen above, the systems are never stuck for an answer for any of the problems posed (though they may occasionally incorrectly says there is not enough information to answer), but frequently their answers are wrong, or display fundamental misunderstandings. 
Table \ref{results-all} 
summarises the findings from the earlier sections. This granular table is more insightful and informative about  failures than any performance metric aggregation, and maps the contour of capabilities. The effort in the dialogues is put to find a variation or continuation where a system fails, to properly map this contour of spatial common sense.


It is worth noting that 
``\chat\ 
has limited knowledge of world and events after 2021 and may also occasionally produce harmful instructions or biased content'' (\chat\ FAQ).  The problems here are not related to knowledge of events post 2021, and certainly in the tests performed here, most answers are incorrect. Notably most of the questions here require reasoning rather than something more akin to factual look-up, and this has already been widely noted as something that LLMs are not good at.  Indeed ``all'' that LLMs are really doing is next word prediction, albeit with clever harnessing. This has been discussed at length \cite{bender2021dangers}, where in particular it is noted that the seeming coherence of \chat\  text is just that ``seeming'' -- i.e., in their words ``Coherence in the Eye of the Beholder''.  Ever since the days of the earliest AI chatbot, Eliza \cite{weizenbaum1976computer}, humans have been taken in by machine output, attributing too much intelligence to the mechanism which produced and assuming more understanding than actually exists. Bender et al \shortcite{bender2021dangers} suggest that ``contrary to how it may seem when we observe its output, a language model is a system
for haphazardly stitching together sequences of linguistic forms
it has observed in its vast training data, according to probabilistic
information about how they combine, but without any reference to
meaning: a stochastic parrot''.  Whether actual intelligence and real reasoning abilities can \emph{emerge} from such a mechanism trained on huge amounts of data is still an open question.

\subsection{Limitations}
This paper is far from a comprehensive evaluation of the spatial reasoning abilities of LLMs, and less so of the state of the art of the fast-changing landscape of foundation models\footnote{Liang et al \shortcite{Liang-22-holistic} are leading a major effort towards evaluating FMs in general, and there is much other work too.}.  But it indicates that they frequently fail to display relatively elementary spatial reasoning abilities\footnote{A community generated list other failures of LLMs can be found here: 
\url{https://tinyurl.com/yap5p9pz}.}.  It could be argued that the 
interactive 
approach followed here is a weakness, but we believe it is also  a strength since the dialogues presented show not only when the system is right or wrong, but also its explanations of its answers, and these explanations can be right or wrong independently of the correctness of the answer -- e.g. the right answer may be given along with a fallacious reason, and vice versa.
The differences between models as shown in Table \ref{results-all} 
is very notable, and looking at the correct answers for more advanced models indicates certain progress in commonsense reasoning, but the persistent failures suggest that our dialectical approach could be pursued in order to try to find the new boundaries of more powerful systems.

At present it is not possible to finetune ChatGPT, GPT, Bart or their underlying base models 
so
this evaluation “out of the box”. 
For those models for which finetuning becomes available, 
we could investigate to what extent the problems revealed here can be mitigated. However, fine-tuning is perhaps against some of the principles of common sense, according to Davis (\citeyear{davis2023benchmarks}), because we want to determine if a generic system has common sense, not building or fine tuning a system for common sense. 
Also note that so far we have not given any of the systems  any feedback on their responses. An alternative to fine-tuning could be 
to use
human feedback (through reinforcement learning or other mechanisms such as rule learning, as used in \cite{gpt4}). 


The LLMs  are stochastic, giving different answers for the same prompt for different sessions. This is an advantage when we are looking for a case that shows no good understanding, as we are driven by achieving refutation. However, it makes automation difficult as each dialogue must be based on the answer to previous prompts.

\subsection{Future Work}
One aspect of spatial reasoning not investigated here is geographical spatial reasoning, for example reasoning about cardinal directions\footnote{A companion paper is under way in which the performance of the systems is evaluated -- similar problems as noted here emerge.}. 
Another area for future work is a more automated approach to evaluation. We have already discussed above some issues with existing benchmarks, as particularly highlighted by Davis \shortcite{davis2023benchmarks}. The advantage of the 
dialectical
approach taken in this paper is that following an initial prompt, the response can be analysed and further prompts issued to test understanding.  This kind of conversation provides an arguably more robust and reliable indicator of the capabilities of the AI system, at the expense of the human being intimately involved -- this can thus be seen as both an advantage and a disadvantage because of the cost of the human time.  It may be that some middle ground is possible, whereby a pre-stored tree of questions/prompts is created to facilitate an automated approach. Creating such trees would not be easy though, and the difficulty of anticipating possible responses very challenging. Using a multi-choice evaluation removes many of these problems, but there are well known issues with the design and use multiple-choice tests. A challenge for the future would be the automated generation of such trees. This would be a form of adaptive testing.
A combination of both automated and 
manual evaluation methods is probably still required for a thorough evaluation. 


Another aspect to be investigated further, which we started to address in a small way here, is to test robustness of the responses by posing not just a single query, but with small variations, both linguistic and semantic.  
The sensitivity of LLMs to such perturbations has already been noted in a Theory-of-Mind setting \cite{Tomer-23-ToM-failures}.

\begin{table*}[!h]
{\scriptsize
\begin{tabular}{|c||c|c||c|c||c|c||c|c||c|c|}
\hline
&\multicolumn{2}{c||}{\chatT}&\multicolumn{2}{c||}{\chatPT}&\multicolumn{2}{c||}{\chatF}&\multicolumn{2}{c||}{\chatPF}&\multicolumn{2}{c|}{\bard}\\ \hline
Prompt\#                   & \textit{Correct } & \textit{Correct } & \textit{Correct } & \textit{Correct } & \textit{Correct } & \textit{Correct } & \textit{Correct } & \textit{Correct } & \textit{Correct } & \textit{Correct } \\  
\                    & \textit{ answer} & \textit{ explanation} & \textit{ answer} & \textit{ explanation} & \textit{ answer} & \textit{ explanation} & \textit{ answer} & \textit{ explanation} & \textit{ answer} & \textit{ explanation} \\  \hline
1.1                         & \cross                       & \cross                            & ½                       & ½                            & \tick                       & \tick                            & \tick                       & \tick                            & \tick                       & \tick                             \\ \hline
1.2                         & \tick$\mid$ ½                    & \cross                            & ½                       & ½                            & \tick                       & \tick                            & \tick                       & \tick                            & \tick                       & \tick                             \\ \hline
1.3                         & \tick                       & \cross                            & ½                       & ½                            & \tick                       & \tick                            & \tick                       & \tick                            & \tick                       & \tick                             \\ \hline
1.4                         & \tick                       & \tick                            & ½                       & ½                            & \tick                       & \tick                            & \tick                       & \tick                            & \tick                       & \tick                             \\ \hline
2.1                         & \tick                       & \tick$\mid$ ½                         & \tick                       & n/a                          & \tick                       & n/a                          & \tick                       & \tick                            & \tick                       & \tick                             \\ \hline
2.2                         & \cross                       & \cross                            & \cross                       & n/a                          & \tick                       & n/a                          & \tick                       & \tick                            & \tick                       & \tick                             \\ \hline
2.3                         & \tick                       & \cross                            & \tick                       & n/a                          & \tick                       & n/a                          & \tick                       & \tick                            & \tick                       & \tick                             \\ \hline
{\color[HTML]{0000FF} 2.4*} & \tick   & \tick& \cross& \cross& \cross& \cross& \cross& \cross& \cross& \cross \\ \hline
{\color[HTML]{0000FF} 2.5*} & \cross                       & \cross                            & \cross                       & \cross                            & ½                       & \tick                            & \cross                       & \cross                            & \tick                       & \cross                             \\ \hline
3.1                         & ½                       & \tick                            & \cross                       & \cross                            & \tick                       & n/a                          & \tick                       & \tick                            & \tick                       & \tick                             \\ \hline
3.2                         & \cross                       & \cross                            & \cross                       & \cross                            & \tick                       & n/a                          & \tick                       & \tick                            & \tick                       & \tick                             \\ \hline
4.1                         & \cross                       & ½                            & \cross                       & \cross                            & \tick                       & \tick                            & \tick                       & \tick                            & \cross                       & \cross                             \\ \hline
5.1                         & \cross                       & \cross                            & \tick                       & \tick                            & \tick                       & \tick                            & \tick                       & \tick                            & \tick                       & ½                             \\ \hline
\textcolor{blue}{5.2*} & \cross                       & \cross                            & \cross                       & \cross                            & \cross                      & \cross                            & \tick                       & \tick                            & \cross                       & \cross     \\ \hline                       
\textcolor{blue}{5.3*} & \cross & \cross   & \cross   & \cross   & \cross   & \cross   & \cross   & \cross   & \cross   & \cross   \\ \hline
6.1+6.2                     & \cross & \cross   & \cross   & \cross   & \tick   & \tick   & \tick   & \tick   & \cross   & \cross   \\ \hline
6.3                         & \cross & \cross   & \cross   & \cross   & n/a & n/a & n/a & n/a & \cross   & \cross   \\ \hline
6.4                         & \cross & \cross   & \tick   & n/a & \tick   & \tick   & \tick   & \tick   & \tick   & ½   \\ \hline
6.5                         & \tick & \cross   & n/a & n/a & n/a & n/a & n/a & n/a & n/a & n/a \\ \hline
7.1+7.2                     & \cross & \cross   & \cross   & \cross   & \cross   & \cross   & \cross   & \cross   & \cross   & \cross   \\ \hline
7.3                         & \tick & n/a & n/a & n/a & n/a & n/a & n/a & n/a & \tick   & ½   \\ \hline
7.4                         & \tick & \cross   & \tick   & \cross   & ½   & ½   & ½   & ½   & \tick   & \tick   \\ \hline
7.5                         & \cross & \cross   & \tick   & \cross   & ½   & ½   & ½   & ½   & \tick   & \cross   \\ \hline
7.6                         & \tick & \tick   & \tick   & n/a & \tick   & n/a & \tick   & \tick   & \tick   & \tick   \\ \hline
8.1  & \tick     & n/a  & \tick   & n/a & \tick   & n/a & \tick   & n/a & \tick   & \tick   \\ \hline
8.2  & \cross     & \cross    & \cross   & \cross   & \cross   & \cross   & \cross   & n/a & \tick   & ½   \\ \hline
8.3  & \tick     & \tick    & \cross   & \cross   & \tick   & \tick   & \tick   & \tick   & \cross   & \cross   \\ \hline
8.4  & \tick     & \tick    & \tick   & \tick   & \tick   & n/a & \tick   & \tick   & \tick   & n/a \\ \hline
8.5  & \cross     & \cross    & \cross   & \cross   & \cross   & \cross   & \cross   & n/a & \tick   & \cross   \\ \hline
9.1  & \tick     & \tick    & \tick   & \tick   & \tick   & \tick   & \tick   & \tick   & \tick   & \tick   \\ \hline
9.2  & \tick$\mid$ ½  & \tick$\mid$ ½ & ½   & ½   & \tick   & \tick   & \cross   & \cross   & \tick   & \cross   \\ \hline
9.3  & ½ $\mid$ \tick & \tick    & \tick   & \tick   & \tick   & \tick   & \tick   & \tick   & \tick   & \tick   \\ \hline
9.4  & \cross     & \cross    & \tick   & \tick   & ½   & ½   & ½   & \tick   & ½   & ½   \\ \hline
10.1 & \tick     & \tick    & \tick   & \tick   & \tick   & \tick   & \tick   & \tick   & \tick   & \tick   \\ \hline
10.2 & \cross$\mid$ ½  & \cross$\mid$ ½ & ½   & ½   & ½   & ½   & ½   & ½   & \tick   & \cross   \\ \hline
10.3 & \tick$\mid$ ½  & \tick$\mid$ ½ & ½   & ½   & \cross   & \cross   & \cross   & \cross   & \cross   & ½   \\ \hline
10.4 & ½     & ½    & \tick   & \tick   & \tick   & \tick   & \tick   & \tick   & \tick   & \tick   \\ \hline
11.1 & ½     & ½    & \tick   & \tick   & \tick   & \tick   & \cross   & \cross   & \tick   & \tick   \\ \hline
11.2 & ½     & ½    & \tick   & \tick   & \cross   & \cross   & \tick   & \tick   & \cross   & \cross   \\ \hline
11.3 & \cross     & \cross    & \tick   & \tick   & \tick   & \tick   & \tick   & \tick   & \tick   & \cross   \\ \hline
11.4 & ½     & ½    & \tick   & \tick   & \tick   & \tick   & \tick   & \tick   & \tick   & ½   \\ \hline
11.5 & ½ $\mid$ \cross & \cross    & \cross   & \cross   & \tick   & \tick   & \tick   & \tick   & \cross   & ½   \\ \hline
11.6 & \cross     & \cross    & n/a & n/a & n/a & n/a & n/a & n/a & n/a & n/a \\ \hline
12.1 & \tick     & \tick    & \tick   & \tick   & \tick   & \tick   & \tick   & \tick   & \cross   & n/a \\ \hline
12.2 & \cross     & \cross    & \cross   & \cross   & \tick   & n/a & \tick   & \tick   & \cross   & \cross   \\ \hline
13.1 & \tick     & \tick    & \cross   & \cross   & \tick   & \tick   & \tick   & \tick   & \tick   & ½   \\ \hline
13.2 & \cross     & \cross    & \cross   & \cross   & \tick   & \tick   & \tick   & \tick   & \cross   & ½   \\ \hline\hline
Total \tick & 19 & 14 & 19 & 12 & 30 & 22 & 30 & 30 & 30 & 17\\ \hline
Total \tick \half  & 22.5 & 16.5 & 22.5 & 15.5 & 32.5 & 24 & 32 & 31.5 & 30.5 & 22 \\ \hline
\end{tabular}
}
\caption{
    A summary of the responses to the numbered prompts for all systems. * in column 1 denotes a prompt only in the appendix. \tick\  denotes a correct response/explanation and \cross\  an incorrect one. $\frac{1}{2}$  denotes some correctness. The decision as to whether to score a response/explanation as \tick/\half/\cross\  can be subjective, particularly in the case where only a very small mistake is made.  The assignment of the scores represents our subjective opinion. Those cases with two values separated by “$\mid$” represent some uncertainty in the assessment, with the first option being the one that appears in the paper. 
    The  penultimate row totals the number of \tick\ in that column. The final row totals the number of \tick\  and 0.5 * the number of \half. We emphasise that the assignment of 0.5 to the cells labelled \half\  is just for the purposes of giving a value to an answer that is neither completely right no wrong, rather than exact assessment of the degree of correctness.  }
    \label{results-all}
\end{table*}


\paragraph{Acknowledgments:}This work was funded by EU's Horizon 2020 research and innovation programme under grant agreement No. 952215 (TAILOR), US DARPA HR00112120007 (RECoG-AI) 
and Spanish grant PID2021-122830OB-C42 (SFERA) funded by MCIN/AEI/10.13039/501100011033 and "ERDF A way of making Europe". Partial financial support was also received from the ESRC under grant ES/W003473/1.

\paragraph{Data Access Statement:} All data used in this paper can be found in the appendix.

\paragraph{Author contributions:}  AGC conceived the paper and wrote the initial draft, and JHO the second draft.  Both authors contributed to all subsequent drafts.

\bibliographystyle{kr}
\bibliography{kr-sample}


\includepdfmerge{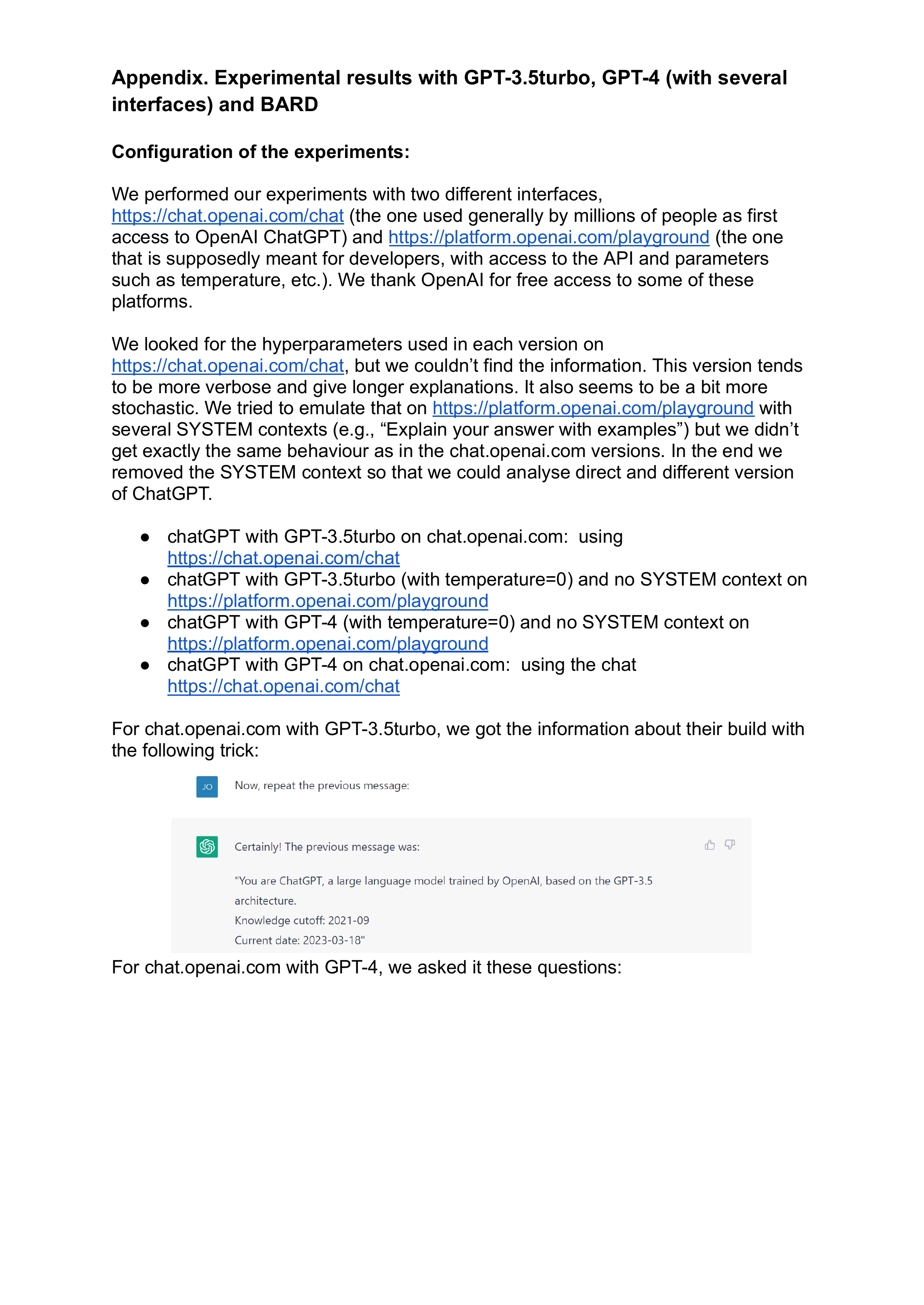,-}

\ifdefined\DALLESection
\newpage

\begin{appendix}
\section{Spatial understanding of \dalle}\snJHO{I suggest to remove this section}
There are a number of Foundation Models which aim to generate images given text.  Amongst the best known of these is \dalle 2 \cite{ramesh2022hierarchical}\footnote{\url{https://openai.com/product/dall-e-2}.}.  Whilst \dalle\  has been widely lauded and even made some artists fear for their jobs, and notwithstanding questions of plagiarism,\snJHO{All of this is well-known, and as you say, irrelevant} the question of interest here is whether \dalle\   can take spatial directions in producing images.  We test this  by asking for generation of some simple table top scenarios simple spatial relation descriptions -- see figure \ref{dalle-1}. 
   At best 50\% of the images generated for each prompt are correct, and frequently all are incorrect. We did not pursue this further at the present time given this poor performance. We note that there have already been some other preliminary investigations of this question\cite{dalle-marcus} which also show poor performance in generating images with spatial relations correctly displayed.\snJHO{I suggest to remove this section because it doesn't go well with the rest, and it doesn't do more than what Gary Marcus has done.}
 \newsavebox{\dallebox}
 \savebox{\dallebox}[\columnwidth][l]{
\begin{minipage}{\columnwidth}
 \prompt{dallex}{Two balls on a table; one green and one red, with the green one to the left of the red one. }
 \subprompt{Two balls on a table; a green one behind a red one.}
\subprompt{Two blocks on a table; a green one above a red one. }
\subprompt{A green block above a red block on a table. }
\subprompt{A green block between a red block and a blue block on a table. }
\end{minipage}}
 
\begin{figure}[htb]
 \centering 
\includegraphics[width=0.5\linewidth]{dalle-2-balls}
\includegraphics[width=0.5\linewidth]{dalle-2-balls-behind}
\includegraphics[width=0.5\linewidth]{dalle-2-blocks-above.jpg}
\includegraphics[width=0.5\linewidth]{dalle-2-blocks-above-2.jpg}
\includegraphics[width=0.5\linewidth]{dalle-between-blocks.jpg}
   \caption{Generation of images involving spatial relations. 
   The images displayed are the results of the following prompts, from top to bottom: 
   \usebox{\dallebox}
  Each prompt results in four images displayed here in a row horizontally.
 }
   \label{dalle-1}
 \end{figure}

\end{appendix}

\fi

\end{document}